\documentclass[journal]{IEEEtran}
\pdfoutput=1
\usepackage{booktabs}
\usepackage{arydshln}
\usepackage[pdftex]{graphicx}
\graphicspath{{../pdf/}{../jpeg/}}
\DeclareGraphicsExtensions{.pdf,.jpeg,.png}
\usepackage[caption=false,font=footnotesize]{subfig}
\begin{document}

\title{Attention-based Convolutional Neural Network \\for Weakly Labeled Human Activities Recognition \\with Wearable Sensors}

\author{Kun~Wang,
		Jun~He,
        and~Lei~Zhang%
\thanks{The work was supported  in part by the National Science Foundation of China under Grant 61203237 and the Industry-Academia Cooperation Innovation Fund Projection of Jiangsu Province under Grant BY2016001-02, and in part by the Foundation of Shanghai Key Laboratory of Navigation and Location-based Services, Shanghai, under Grant 20040.\textit{ (Corresponding author: Lei Zhang.)}}% 
\thanks{Kun Wang and Lei Zhang are with School of Electrical and Automation Engineering, Nanjing Normal University, Nanjing, 210023, China (e-mail: iskenn7@gmal.com, leizhang@njnu.edu.cn). }% 
\thanks{Jun He is with the School of Electronic and Information Engineering,Nanjing University of Information Science and Technology, Nanjing, 210044, China (e-mail: jhe@nuist.edu.cn). }% 
\thanks{Kun Wang and Jun He contribute equally to the article.}
}

\maketitle

\begin{abstract}
Unlike images or videos data which can be easily labeled by human being, sensor data annotation is a time-consuming process. However, traditional methods of human activity recognition require a large amount of such strictly labeled data for training classifiers. In this paper, we present an attention-based convolutional neural network for human recognition from weakly labeled data. The proposed attention model can focus on labeled activity among a long sequence of sensor data, and while filter out a large amount of background noise signals. In experiment on the weakly labeled dataset, we show that our attention model outperforms classical deep learning methods in accuracy. Besides, we determine the specific locations of the labeled activity in a long sequence of weakly labeled data by converting the compatibility score which is generated from attention model to compatibility density. Our method greatly facilitates the process of sensor data annotation, and makes data collection more easy.
\end{abstract}


\IEEEpeerreviewmaketitle

\section{Introduction}
\IEEEPARstart{I}{n} the recent years, the wide diffusion of mobile devices has made human activity recognition(HAR) based on wearable sensors\cite{pei2013human} become a new research point in the field of artificial intelligence and pattern recognition\cite{foerster1999detection}\cite{mannini2010machine}, and there are prevailing applications which benefit from HAR include sports activity detection\cite{yang2015deep}, smart homes\cite{rashidi2009keeping} and health support\cite{magherini2013using}, et al. Those sensors such as accelerometers, gyroscopes and magnetometers\cite{cornacchia2017survey}, which are embedded on mobile devices, can generate time-series data for HAR. But unlike videos or images which can be smoothly annotated by humans, it is laborious to accurately segment a specific type of activity from a long sequence of sensor data. Nevertheless, it is feasible for data collector to identify whether a labeled activity takes place in a long recorded sensor data. This kind of data is corresponding to the “weakly labeled sensor data”. Consequently, the new challenge is whether we can exploit the weakly labeled sensor data to accurately predict activities and determine the exact location of the labeled activity. 
\\
\indent
Traditional methods, which have been developed to facilitate human activity recognition, are inside the range of supervised learning. For instance, the pervious methods include SVM\cite{anguita2012human} and Random Forest\cite{peterek2014comparison}, which require to extract handcrafted features as the inputs of classifiers. Later, deep learning, and in particular, convolutional neural networks, has been diffusely used in the field of HAR. Although deep learning models have excellent performance in HAR, there are some challenges need to be addressed, the main one of which is the need for labeled datasets for ground truth annotation\cite{cruciani2018automatic}.
\\
\indent
Recently, an end-to-end-trainable attention module for convolutional neural network architectures built for image classification has been proposed\cite{jetley2018learn}. Inspired by that, we propose a new weakly supervised learning method of human activity recognition, which can be trained by weakly labeled sensor data and determine the location of the labeled activity.
\\
\indent
In this paper, we exploit the attention-based convolutional neural network model to recognize activities from sensor data with weak labels. The dataset only obtains the information about what activity occurred in a sequence of sensor data without knowing the specific time of this activity. Besides, the precise locations of the specific labeled activity can be determined by utilizing the mechanism of attention.
\\
\indent
The remainder of this paper is structured as follows: Section \uppercase\expandafter{\romannumeral2} summarizes related work of human activity recognition. In section \uppercase\expandafter{\romannumeral3}, we propose our attention model for HAR. Section \uppercase\expandafter{\romannumeral4} presents and examines the experimental results. In section \uppercase\expandafter{\romannumeral5}, we draw out our conclusion.

\section{Related Works}
\indent The pervious methods used to extract features manually. For example, Bao and Intille\cite{bao2004activity} corroborated that accelerometer sensor data is appropriate for activities recognition. They extracted handcrafted features (mean, energy, frequency and domain entropy) from accelerometer data and then fed these features into different classifiers: decision table\cite{witten1999weka}, K-nearest neighbor(KNN)\cite{pirttikangas2006feature}, decision tree\cite{todorovski2003combining}, and Nave Bayes\cite{witten1999weka}. Kwapisz et al.\cite{kwapisz2011activity} also manually extracted features (i.e. average and standard deviation) from the accelerometer sensor of smartphones, and recognized six different activities using classifiers such as decision trees(J48), multilayer perceptions(MLP), and logistic regression. However, these methods based on feature engineering\cite{kao2009development} had low performance in distinguishing similar activities such as walking upstairs and walking downstairs\cite{ronao2015deep}. Besides, it is difficult in both choosing suitable features and extracting features from sensor data manually. 
\\
\indent Therefore, another increasing line of research in human activity recognition based on wearable sensors aims at avoiding the design of handcrafted features, operation that requires human work and expert knowledge\cite{jordao2018human}, these works employ rapid developing deep learning\cite{lecun2015deep}, especially convolutional neural networks(ConvNet), to learn the features and the classifier simultaneously\cite{jordao2018human}. For example, Chen and Xue\cite{chen2015deep} fed raw signal into a sophisticated ConvNet, which had an architecture composed of three convolutional layers and three max-pooling layers. Furthermore, Jiang and Yin\cite{jiang2015human} converted the raw sensor signal into 2D signal image by utilizing the techniques of a specific permutation algorithm and discrete cosine transformation(DCT), then fed the 2D signal image into a two layer 2D ConvNet to classify this signal image equaling to the desired activity recognition. Ordez et al.\cite{ordonez2016deep} proposed an architecture comprised of convolutional and LSTM recurrent units(DeepConvLSTM), which outperforms CNN. However, these methods all belong to supervised learning\cite{shoaib2015survey}, that is to say that all methods require massive data with perfect ground-truth for training. 

\begin{figure}[!t]
	\centering
	\includegraphics[width=3.5in]{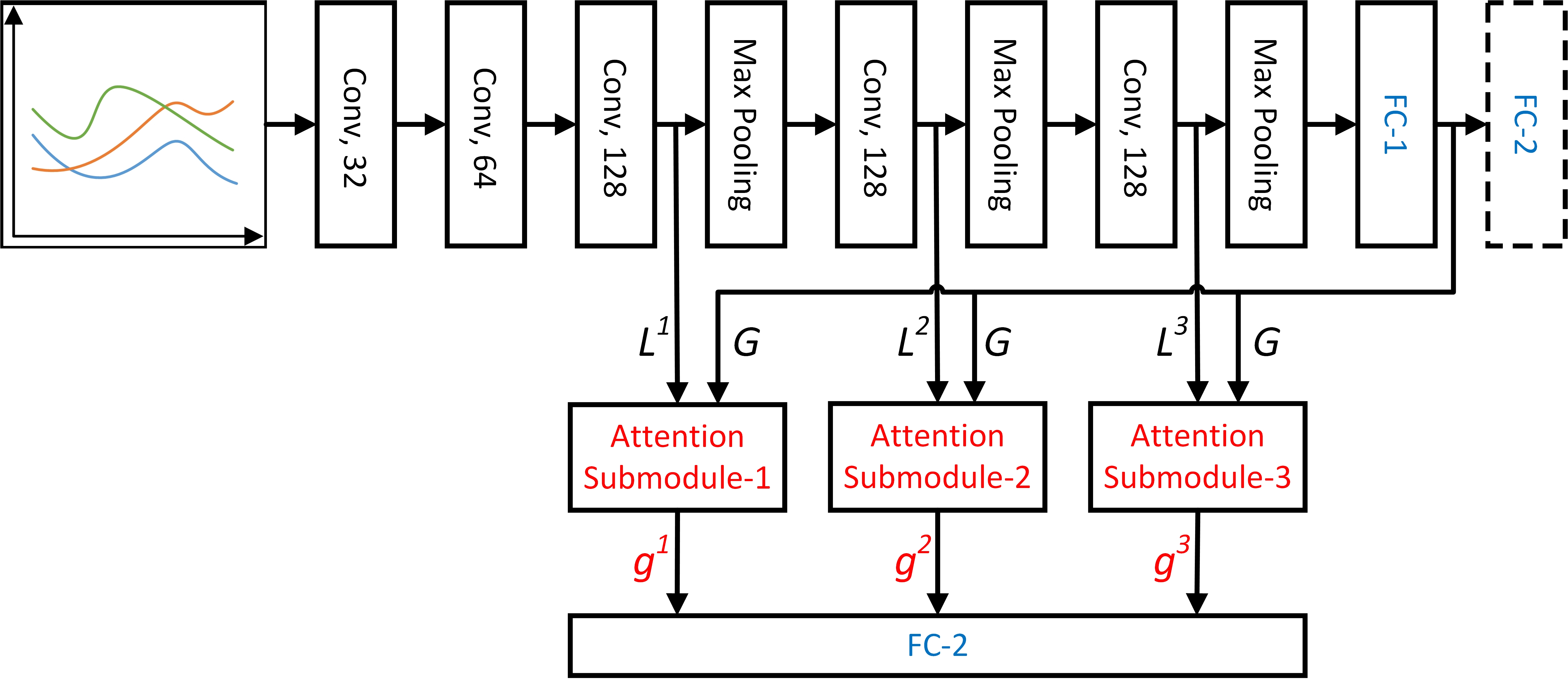}
	\caption{The architecture of our model}
	\label{fig:fig1}
\end{figure}

\section{Model}
\indent
The aim of our model is to recognize and locate human activity in weakly labeled sensor data. The model consists of fundamental CNN pipelines and attention submodule, as the Fig. 1 shows. The advantage of our attention mechanisms is that it can identify salient activity data areas and enhance their influence, while likewise suppressing the irrelevant and potentially confusing information in other activity data areas. Hence, it is very applicable to weakly supervised learning. The model is able to focus on labeled activity among a long sequence of sensor data, and while filter out a large amount of background noise signals. Besides, we propose a novel method to convert compatibility score to compatibility density which is used for locating, because the structure of the weakly labeled sensor data is quite different from the image data. 

\subsection{Attention Submodule}
\indent
The attention method mainly focuses on implementing a compatibility computation between local feature vectors extracted at intermediate layers in the CNN pipeline and the global feature vector that used to be fed into the linear classification layers at the end of the pipeline\cite{jetley2018learn}, which is illustrated in Fig. 1. We denote by $L^s=\left\lbrace l_1^s,l_2^s,…,l_n^s\right\rbrace $ the set of feature vectors extracted at a given convolutional layer $s\in\left\lbrace 1,2,…,S\right\rbrace$, where $l_i^s$ is the $i-th$ feature vector of n total spatial locations in the local feature vector $L^s$. The global feature vectors $G$ generated by input data that passes through entire CNN pipelines, which used to be fed into final fully connected layers to produce the results of classification, now are combined with local feature vectors $L^s$ by a compatibility function C. The compatibility function C can be defined in various ways\cite{bahdanau2014neural}\cite{xu2015show}, we can concatenate the $G$ and $l_i^s$ by an addition operation and then use a dot product between a weight vector $u$ and $l_i^s+G$:
\begin{equation}
c_i^s=\left\langle u,l_i^s+G\right\rangle ,i\in\left\lbrace 1,…,n\right\rbrace
\end{equation}
where $c_i^s$ is the compatibility score and the weight vector $u$ here can play a role in learning the universal set of features relevant to the activity categories in the sensor dataset. In addition, we can also use the dot product between $G$ and $l_i^s$ to weigh up their compatibility:
\begin{equation}
c_i^s=\left\langle l_i^s,G\right\rangle,i\in\left\lbrace 1,…,n\right\rbrace
\end{equation}
\\
\indent After the computing process, we have a set of compatibility score $C(L^s,G)=\left\lbrace c_1^s,c_1^s,…,c_n^s \right\rbrace$, which are then normalized to $A^s=\left\lbrace a_1^s,a_2^s,…,a_n^s \right\rbrace$ by a softmax function:
\begin{equation}
a_i^s=\frac{\exp(c_i^s)}{\sum_{j}^{n}\exp(c_j^s)}
\end{equation}
or a tanh function:
\begin{equation}
a_i^s=\frac{\exp(c_i^s)-\exp(-c_i^s)}{\exp(c_i^s)+\exp(-c_i^s)}
\end{equation}
\\\indent The normalized compatibility scores $A^s$ are used to produce a single vector $g^s$ for each layer $s$ by element-wise weighted averaging:
\begin{equation}
g^s=\sum\limits_{i=1}^{n}a_i^s \cdot l_i^s 
\end{equation}
\\
\indent Most crucially, the global feature descriptor $G$ of the input data now can be replaced by the $g^s$. The new global vectors $g^s$ are concatenated into a single vector $g=\left[ g^1,g^2,…,g^s\right] $ and then fed as input to the linear classification step.

\subsection{Fundamental CNN}
The fundamental model that the attention submodules based on is a deep CNN, which comprises convolutional, pooling and fully-connected layers. As depicted in Fig. 1, the shorthand description is: $C(32)-C(64)-C(128)-P-C(128)-P-C(128)-P-FC(128)-softmax$, where $C(L^s)$ denotes a convolutional layer $s$ with $L^s$ feature maps, $P$ a max pooling layer, $FC(n)$ a fully connected layer with $n$ units and $softmax$ a softmax classifier. Besides, the output of each layer is transformed using a ReLU activation function. The aim that there is no pooling layer following the first two convolutional layer is to ensure that the local feature maps extracted from three convolutional layers $C(128)$ which are utilized for estimating attention have a higher resolution. We deliberately design the same dimensionality between the local features $L^s$ and the global features $G$ (both are 128) in order to avoid mapping the different dimensionality of feature vectors to the same one which can cause extra computational cost. 

\subsection{Location Function}
For an original CNN architecture, a global feature descriptor $G$ usually is computed from the input data and passed through a fully connected layer to output class prediction probabilities. In order to render the classes linearly separable from one another, the $G$ must be expressed by mapping the input into a high-dimensional space where salient higher-order data concepts are represented by different dimensions. The attention method put the filters at the earlier stage in the CNN pipeline to learn similar mappings, which is compatible with the one that outputs $G$ in the original architecture\cite{jetley2018learn}. Due to the attention mechanism of our model, the compatibility score $C(L^s,G)$ should be high if and only if the correspond patch contains parts of the dominant data category. That is to say, for weakly labeled sensor data, the areas with high compatibility score are inclined to be the locations where the labeled activity occurs, while the compatibility scores of those areas where background activity takes place are low. Taking advantage of this notion, we can not only get a good recognition results, but also determine the specific location of the labeled activity in a long sequence of weakly labeled data.

Assume that we have a set of compatibility score $C(L^s,G)=\left\lbrace c_1^s,c_2^s,…,c_n^s \right\rbrace$, where $c_i^s$ is the specific compatibility score of the $i-th$ spatial location of $n$ total spatial locations in the layer $s$. Then a varied width slide window is used to sum up the score within the section at each location of the compatibility score:
\begin{equation}
s_i=\left\lbrace \begin{array}{rcl}
\sum\limits_{j=1}^{i+\frac{w}{2}}c_j^s & \mbox{for} & i<\frac{w}{2}
\\
\sum\limits_{j=i-\frac{w}{2}}^{i+\frac{w}{2}}c_j^s & \mbox{for} & \frac{w}{2}\leq i\leq n-\frac{w}{2}
\\
\sum\limits_{j=i-\frac{w}{2}}^{n}c_j^s & \mbox{for} & i>n-\frac{w}{2}
\end{array}\right. 
\end{equation}
where the location score $s_i$ is corresponding to the density of the compatibility score around the spatial location $i$, the range of this calculation is equal to the slide window width which is varied as the spatial location i changes (the maximum is $w$). The total spatial location $n$ is equal to the length of the set of compatibility score $C(L^s,G)$. Fig. 2 illustrates the above operation process and now we have a set of compatibility density $D=\left\lbrace d_1,d_2,…,d_n\right\rbrace $. The peak points of compatibility density curve denote that these points are most likely where the labeled activity happen. Since activities cannot take place at just one point but a section of data, the areas around peak points are also regarded as the section of the labeled activity. Besides, the coverage of these areas should be defined by the width of slide window. For example, if spatial location $i$ is the peak point of the compatibility density $D$ curve, the section $[i-\frac{w}{2},i+\frac{w}{2}]$ is considered to be the activity area.
\begin{figure}[!t]
	\centering
	\includegraphics[width=3in]{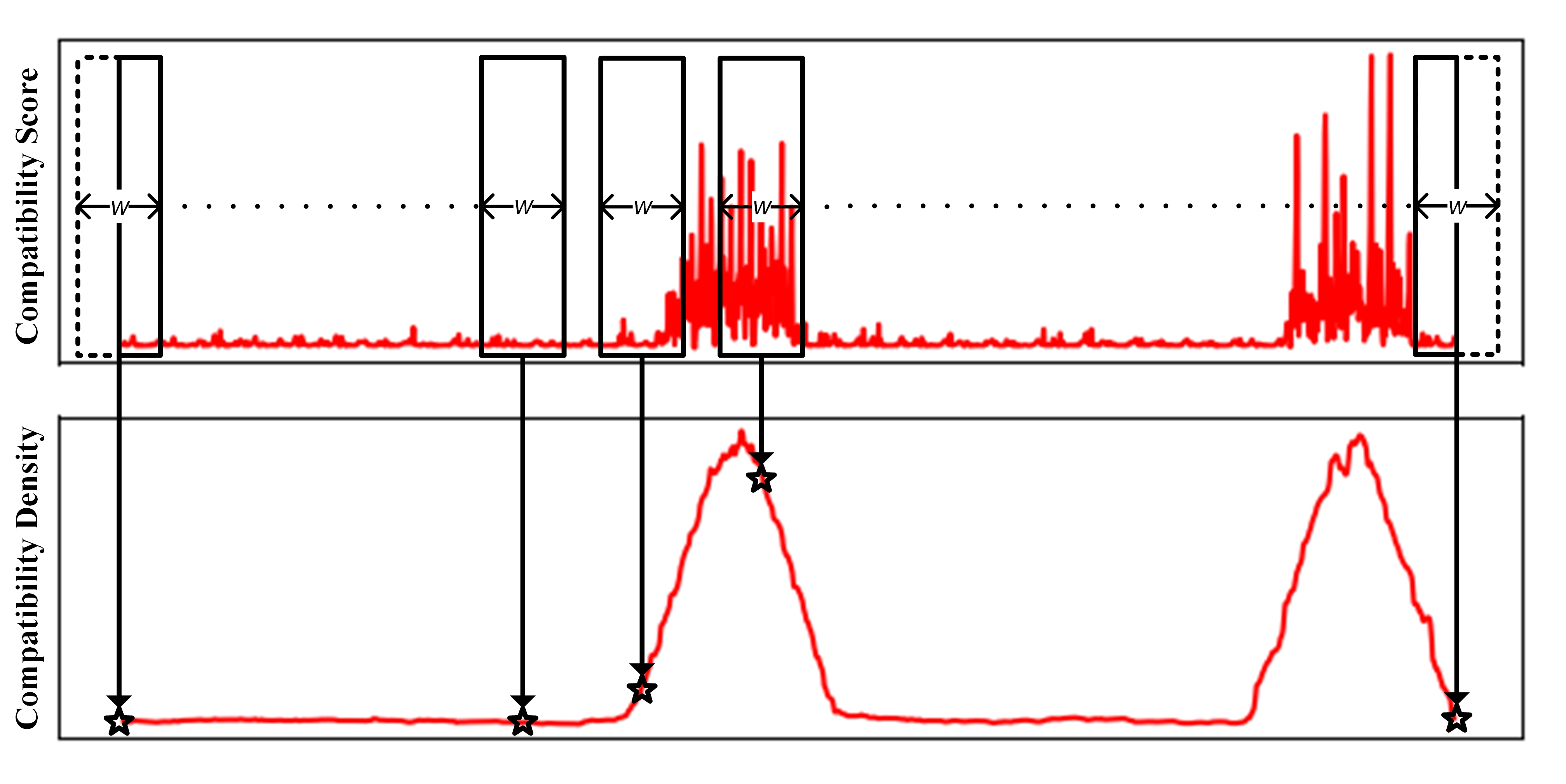}
	\caption{Process of conversion from compatibility score to compatibility density}
	\label{fig:fig2}
\end{figure}

\section{Experiments}
\indent
The experiments are composed of three parts. Firstly, we validate whether our attention model has the capacity to implement traditional human activity recognition. We chose to use the public dataset Human Activity Recognition Using Smartphones Dataset (UCI HAR Dataset). Secondly, we explore the performance of weakly supervised human activity recognition with the weakly labeled sensor dataset. Both types of above experiments used a simple CNN model and the DeepConvLSTM model proposed in \cite{ordonez2016deep} as the baseline models. Among them, the CNN baseline model consists of four convolutional layers with 64 convolution filters and two fully connected layers with 128 units, and then outputs the classification result by the Softmax layer. We use the default parameter setting for DeepConvLSTM according to \cite{ordonez2016deep}. Thirdly, we investigate the location performance of the attention model. The experiments are performed on a workstation with CPU AMD Ryzen 5, 8 GB memory, and a NVIDIA GPU 1050 with 5GB memory. All algorithm is implemented in Python by using the machine learning framework TensorFlow. In the experiments, the number of epoch was set to 100 and Adam optimization method was used to train our model. The learning rate was set to 0.001 and the input batch size was 50. 
\\\indent
We refer to the network Net with attention at the last level as \textit{Net-att}, at the last two levels as \textit{Net-att2}, and at the last three levels as \textit{Net-att3}. We use \textit{dot} to denote the operation of the dot product for matching the local and global features and use \textit{pc} to denote the operation of parametrized compatibility. We denote by \textit{sm} the utilization of softmax normalization and by \textit{tanh} the utilization of tanh normalization.
\subsection{Dataset}
\textit{1) UCI HAR Dataset}\cite{anguita2013public}: This dataset consists of 12 daily activities, namely 3 static activities (standing, sitting, lying), 3 dynamic activities (walking, going upstairs, going downstairs), and the switch of 3 static activities (standing-sitting, sitting-standing, standing-lying, lying-sitting, standing-lying, lying-standing). The data are recorded from a Samsung Galaxy smartphone, which collected three-axial linear acceleration and three-axial angular velocity at a constant rate of 50 Hz using its embedded accelerometer and gyroscope. The collectors utilized the video-recorded to label the data manually, and then divided the data into two sets randomly, where 70\% was selected to generate the training data and 30\% the test data. The sensor data was then pre-processed by applying a noise filter and then sampled in a fixed width sliding window of 2.56 seconds and 50\% overlap (window width is 128). We utilize six activities of this dataset including three static and three dynamic activities for our experiments.

\textit{2) Weakly Labeled Dataset}: The dataset includes five kinds of activities:”walking”, ”jogging”, ”jumping”, ”going upstairs” and ”going downstairs”. The need to pay attention to is that “walking” is the background activity for four other kinds, as illustrated in the Fig. 3. This dataset is collected from 3-axis acceleration of iPhone, which was placed in 7 participants’ right trouser pocket. The smartphone accelerometer has a sampling rate of 50Hz. We divide raw data by distinguishing different participants, then use a fixed width sliding window of 40.96 seconds (window width is 2048) to sample the data. Finally, this collected weakly labeled dataset consists of 76,157 sequences of data, 70\% of which is used for training, 10\% for validating, and 20\% for testing. The statistics of different activity samples are shown in Table \uppercase\expandafter{\romannumeral1}.

\begin{figure}[!t]
	\centering
	\includegraphics[width=3in]{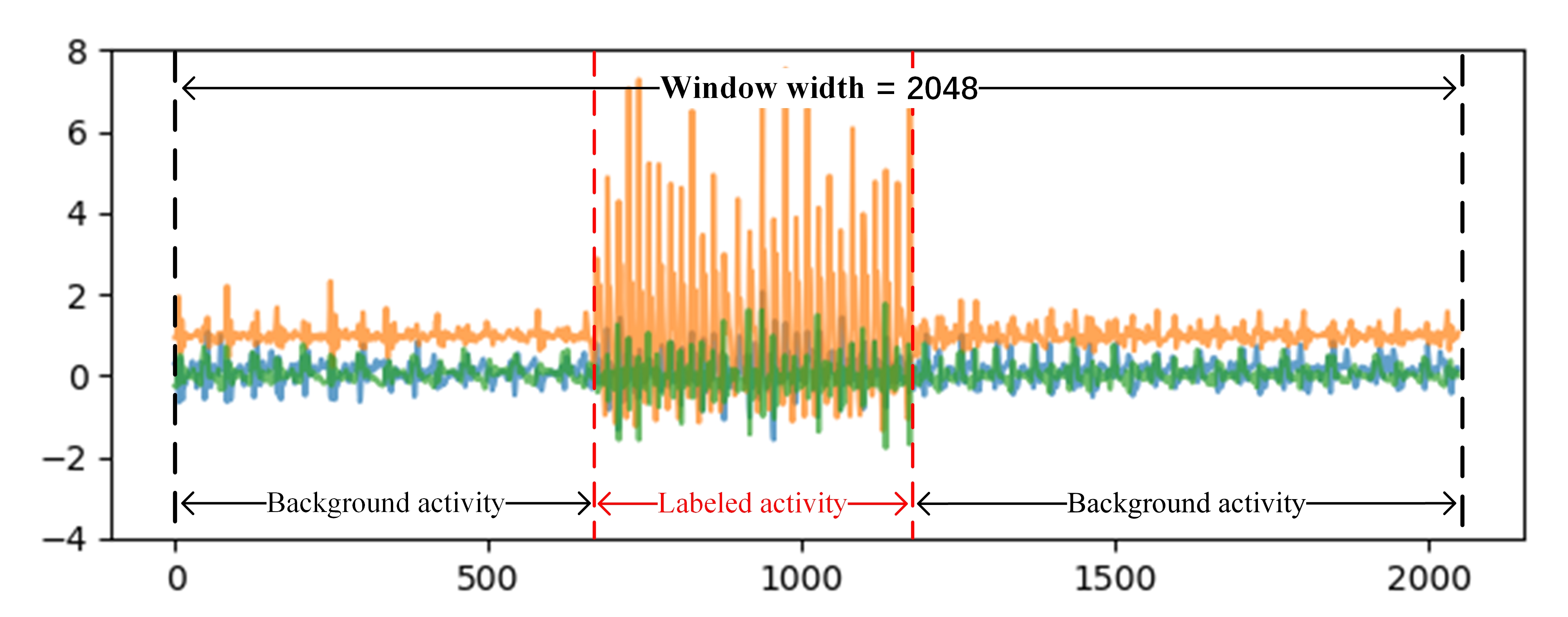}
	\caption{Weakly labeled sample data, the labeled activity is "going upstairs" and the background activity is "walking"}
	\label{fig:f3}
\end{figure}

\begin{table}[!t]
	\renewcommand{\arraystretch}{1.3}
	\caption{Weakly Labeled Dataset Statistics}
	\label{t1}
	\centering
	\begin{tabular}{|c|c|c|}
		\hline
		\bfseries Activity & \bfseries Label & \bfseries Number\\
		\hline
		going upstairs & 0& 20194\\
		\hline
		going downstairs & 1& 18611\\
		\hline
		jumping & 2& 14088\\
		\hline
		jogging & 3& 23264\\
		\hline
		total & -& 76157\\
		\hline
	\end{tabular}
\end{table}

\subsection{Experiments on UCI HAR Dataset}
We compared the experimental results of our attention model, the fundamental CNN model, and two classical deep learning models\cite{ordonez2016deep} on the UCI HAR Dataset in the metrics of classification accuracy. The computational cost that is defined as the average number of predicted sequences per second is used to evaluate the efficiency. The results are shown in Table \uppercase\expandafter{\romannumeral2}. On this dataset, all of the models perform comparably well, which indicates that our model also performs well on traditional supervised training tasks. Regarding the efficiency, our approach can predict more than 5000 sequences per second which is relate slow than CNN but comfortably ahead of the DeepConvLSTM model.
\begin{table}[!t]
	\renewcommand{\arraystretch}{1.3}
	\caption{Experiment on UCI HAR Dataset}
	\label{t2}
	\centering
	\begin{tabular}{lll}
 \toprule[1.2pt]
  Model &  Accuracy& Efficiency\\
 \midrule
 \textbf{-Architectures without attention-}& &\\
 The fundamental CNN & 93.16\% &8046 seqs/s\\
 CNN\cite{ordonez2016deep}& 93.21\%&8745 seqs/s\\
 DeepConvLSTM\cite{ordonez2016deep}& 93.54\%&580 seqs/s\\
 \midrule
 \textbf{-Architectures with attention-}& &\\
 Net-att2-dot-sm& 92.27\%&6573 seqs/s\\
 \cdashline{1-3}[0.8pt/2pt]
 Net-att-pc-tanh& 93.21\%&6280 seqs/s\\
 Net-att2-pc-tanh& 93.41\%&5953 seqs/s\\
 Net-att3-pc-tanh& 92.70\%&5517 seqs/s\\
 \bottomrule[1.2pt]
	\end{tabular}
\end{table}
\subsection{Experiments on Weakly Labeled Dataset}
We compare our model to the fundamental CNN model to study whether our methods can better recognize activities from weakly labeled dataset. Besides, we use two classical deep learning models\cite{ordonez2016deep} as baseline model. The proposed attention model produce visible performance improvement over the fundamental model under certain conditions, as shown in Table \uppercase\expandafter{\romannumeral3}. Specifically, the \textit{Net-att3-pc-tanh} model achieves a 3.65\%, 4.21\% and 3.79\% improvement over the fundamental CNN model, the classical CNN model and the DeepConvLSTM model for weakly labeled data recognition. As is evident from Fig. 4, the utilization of attention mechanism facilitates the model to focus on the areas of labeled activity supposed to be extracted features while ignoring the background activity information. In terms of efficiency, compared with the classical CNN model, our model can achieve greater performance without increasing unduly computational burden.
\begin{table}[h]
	\renewcommand{\arraystretch}{1.3}
	\caption{Experiment on Weakly Labeled Dataset}
	\label{t2}
	\centering
	\begin{tabular}{lll}
		\toprule[1.2pt]
		Model &  Accuracy & Efficiency\\
		\midrule
		\textbf{-Architectures without attention-}& &\\
		The fundamental CNN & 90.18\%& 1145 seqs/s\\
		CNN\cite{ordonez2016deep}& 89.62\%& 1042 seqs/s\\
		DeepConvLSTM\cite{ordonez2016deep}& 90.04\%& 174 seqs/s\\
		\midrule
		\textbf{-Architectures with attention-}& &\\
		Net-att3-dot-sm& 84.15\%& 799 seqs/s\\
		\cdashline{1-3}[0.8pt/2pt]
		Net-att3-dot-tanh& 91.58\%& 801 seqs/s\\
		\cdashline{1-3}[0.8pt/2pt]
		Net-att3-pc-sm& 85.27\%& 773 seqs/s\\
		\cdashline{1-3}[0.8pt/2pt]
		Net-att-pc-tanh& 92.84\%& 976 seqs/s\\
		Net-att2-pc-tanh& 93.55\%& 895 seqs/s\\
		Net-att3-pc-tanh& 93.83\%& 775 seqs/s\\
		\bottomrule[1.2pt]
	\end{tabular}
\end{table}
\begin{figure}[h]
	\centering
	\subfloat[Att3-pc-tanh]{\includegraphics[width=3in]{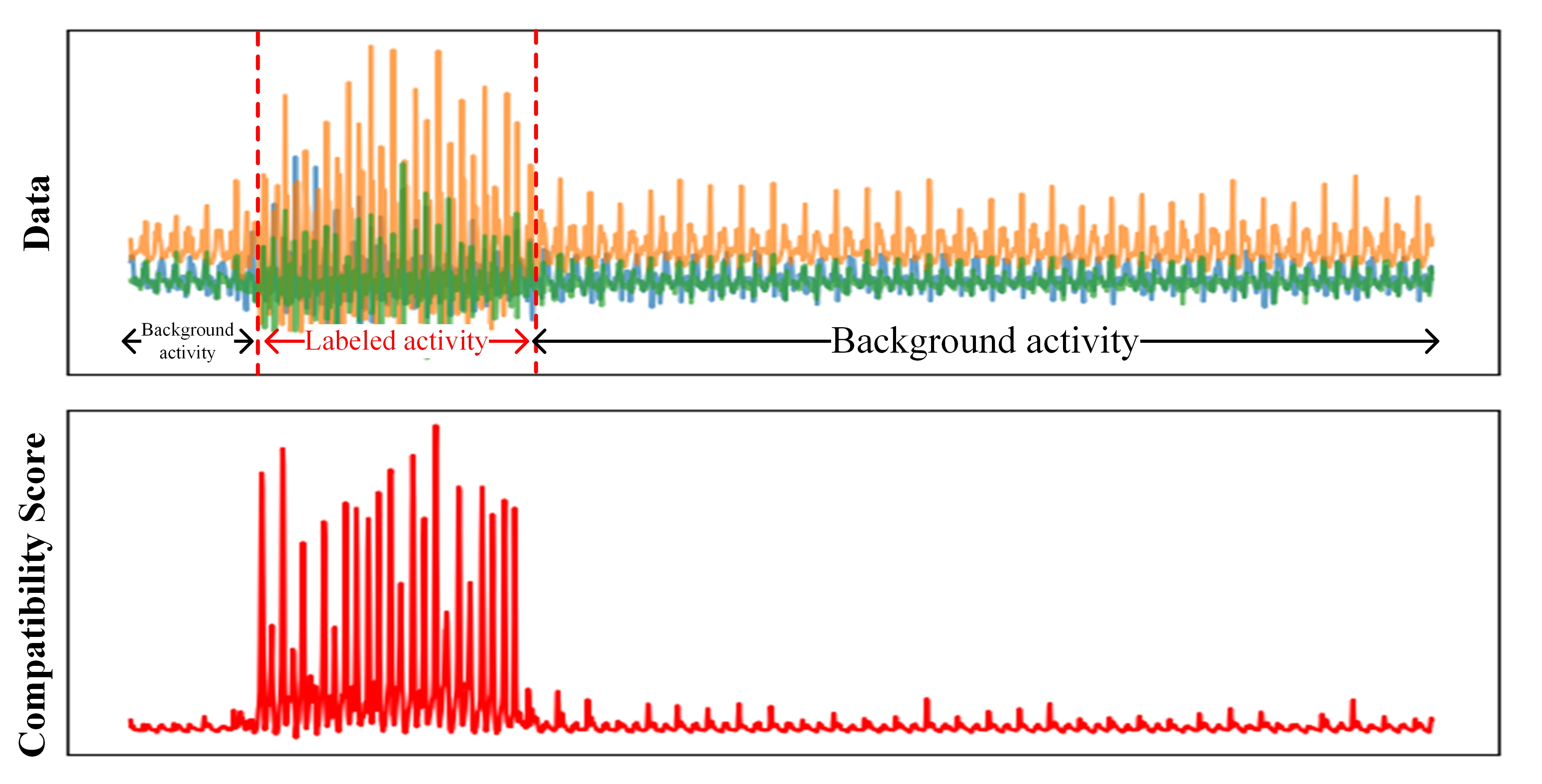}
	\label{f12-1}}
	\hfil
	\subfloat[Att3-dot-sm]{\includegraphics[width=3in]{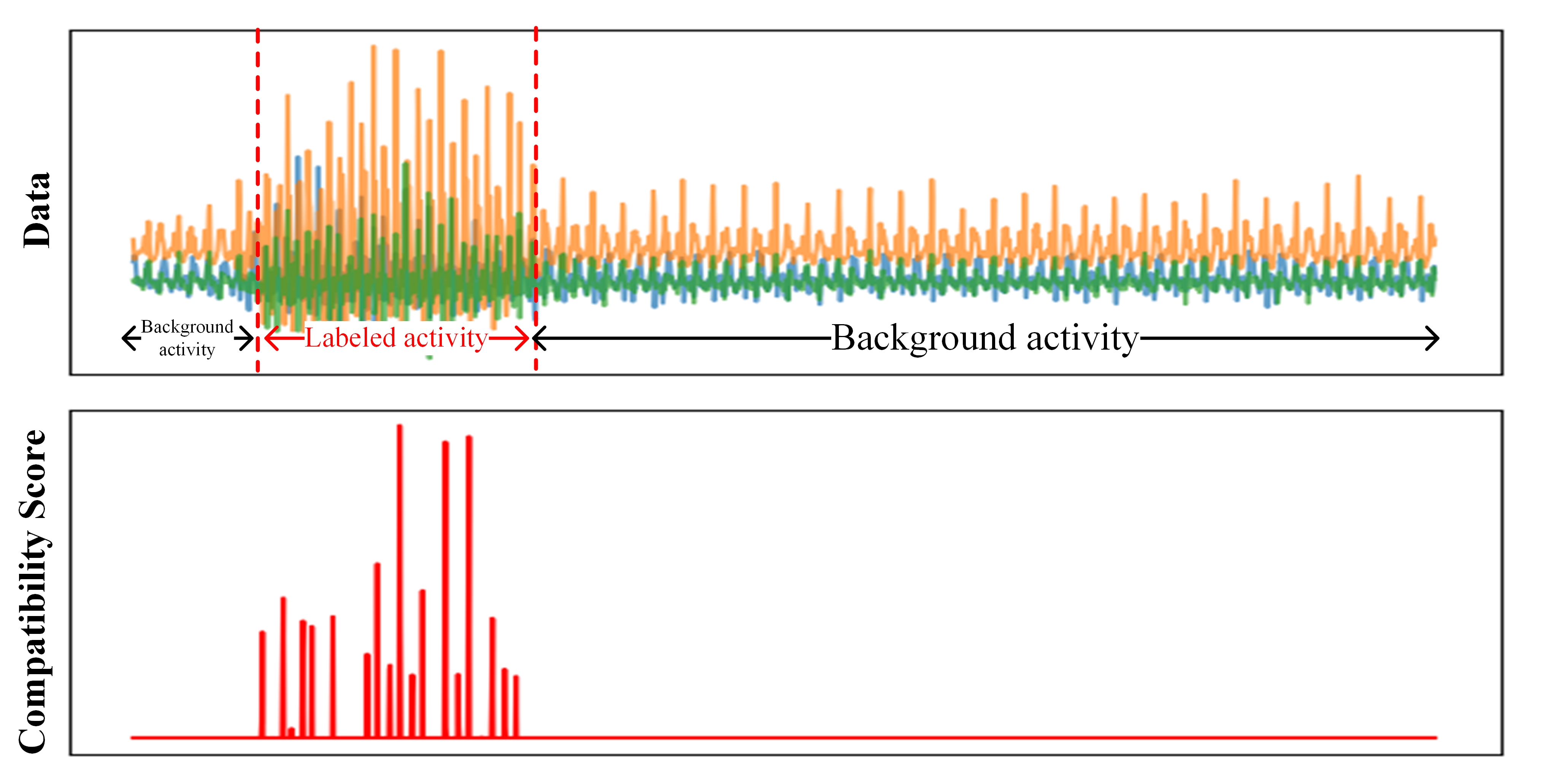}
	\label{f12-2}}
	\caption{Illustration of the weakly sensor data sample and the generated compatibility score}
\end{figure}
\\
\indent In contrast to the small difference in recognition performance among the attention models on UCI HAR dataset, that the performance on weakly labeled dataset is relatively high. The combination of the dot product operation and the softmax normalization is implemented on the weakly labeled dataset that consist of long sequences data generates the abnormal compatibility score. As Fig. 4(b) shows,there are only a few points having excessively effective value, which results in the feature vectors produced by combining the local feature vectors and the defective compatibility score that lack crucial information beside of these valid points. We address the above problem by employing \textit{Att-pc-tanh} architecture, where the operation of parametrized compatibility can generate a set of compatibility scores with relatively low disparity and meanwhile the compatibility scores are not jointly normalized by the softmax operation but are normalized independently using pointwise tanh function. 

\subsection{Location Experiments}
We conduct the location experiment using the proposed attention \textit{Net-att3-pc-tanh} model on the weakly labeled data sequences which have manually labeled ground truth by matching the video and the sensor data. Converting the compatibility score to the compatibility density provides more clarity in determining the locations of labeled activity, because compared with the peak points of the compatibility score curve, that of the compatibility density curve concentrate on the points where the labeled activity happen more intensively. We mark the locations by utilizing windows whose width are equal to that of the slide window used for conversion of scores. In our experiment, the width of the slide window is set to 128. The center of these windows are determined by the local maximum points of the compatibility density curve instead of the compatibility score. The location windows depend on compatibility density locate within the margin of labeled activity as expected, while another one cross the boundary, as is illustrated between Fig. 5(a) and Fig. 5(b).
\begin{figure}[!t]
	\centering
	\subfloat[Labeled activity location based on compatibility density]{\includegraphics[width=3in]{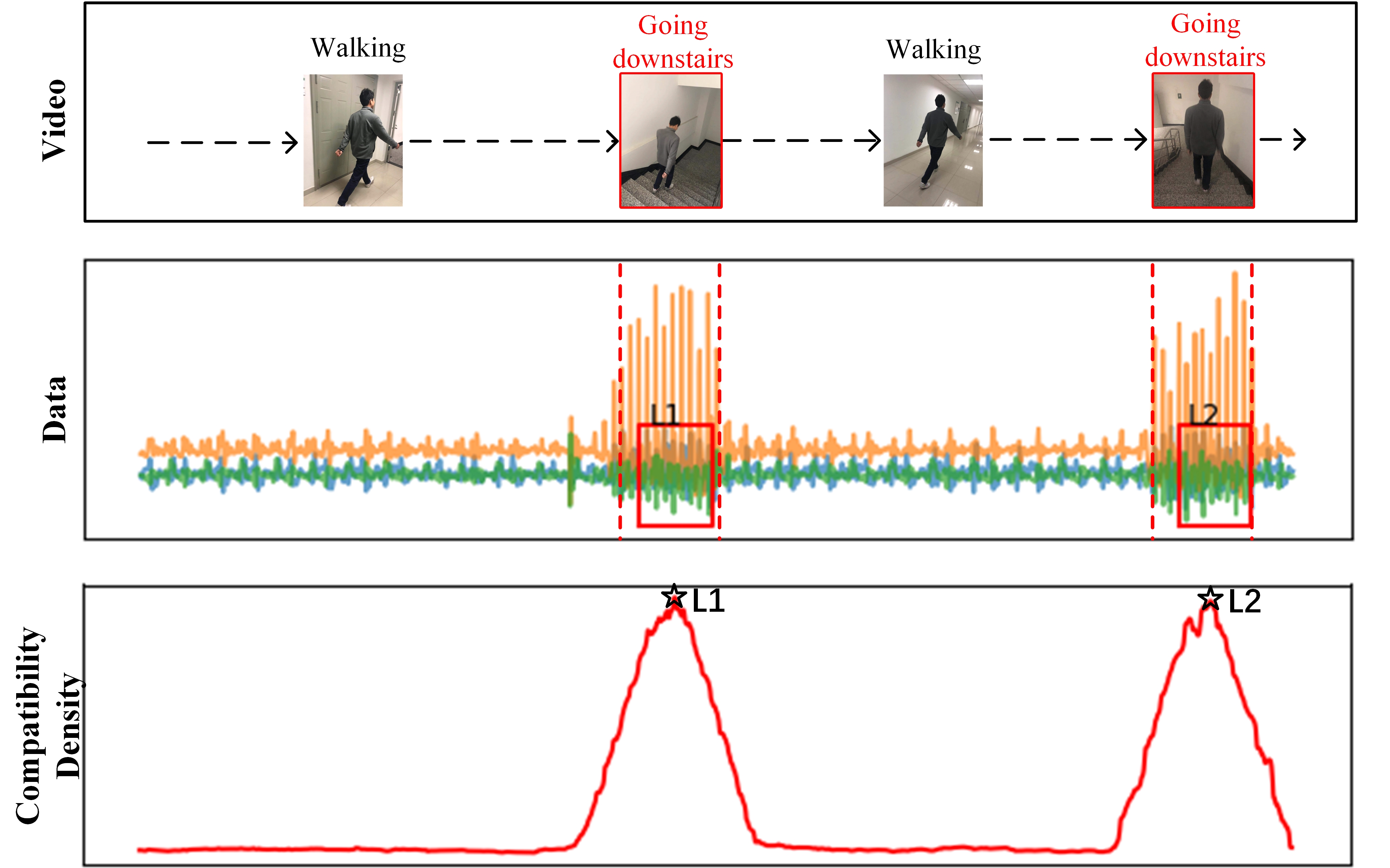}
		\label{Labeled activity location based on compatibility density}}
	\hfil
	\subfloat[Labeled activity location based on compatibility score]{\includegraphics[width=3in]{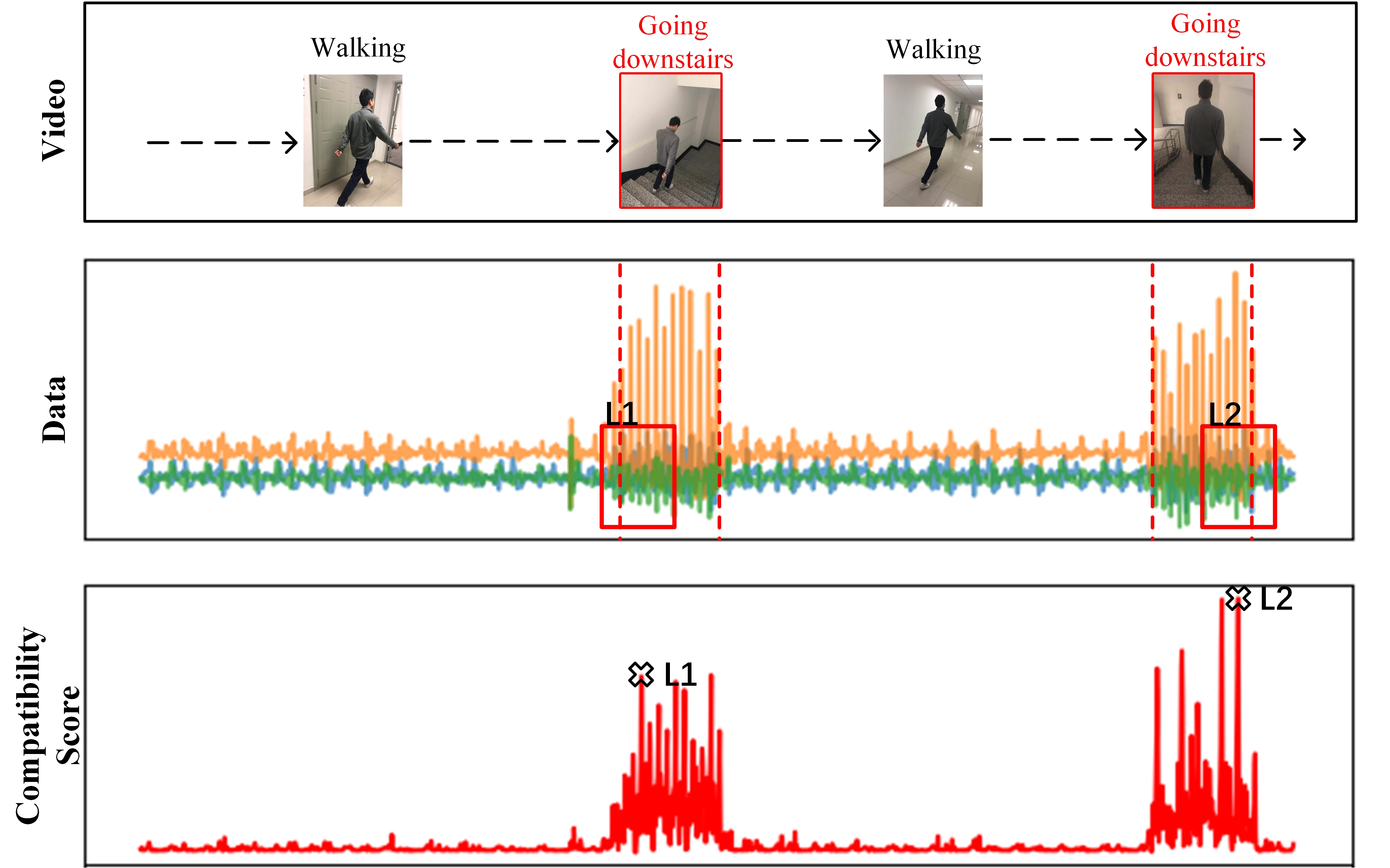}
		\label{Labeled activity location based on compatibility score}}
	\caption{Comparison of the location performance between the compatibility density and the compatibility score}
	\label{fig_sim}
\end{figure}

\section{Conclusion}
We study the attention-based convolutional neural network for weakly labeled human activities recognition with wearable sensors. The experimental results demonstrate that the proposed attention model is comparable with the classical methods of human activity recognition on traditional UCI HAR Dataset. Crucially, our model can achieve significant accuracy improvement on the weakly labeled dataset, compared to the classical deep learning models. We discuss how attention model achieves higher accuracy and why the \textit{Att-pc-tanh} architecture perform well than the \textit{Att-dot-sm}. 
\\
\indent
Usually, it is hard for data recorder to keep a fixed activity, and the weakly labeled sensor data inevitably occurs in the data collection process. Our attention-based CNN method can greatly facilitate the data collection process based on wearable sensor data. The proposed model can utilize the weakly labeled data to avoid the time-consuming process of traditional data annotation, and do not increase unduly computational cost compared to the classical CNN model. Taking advantage of the mechanism of attention, we utilize the compatibility score by converting it to compatibility density to determine the specific location of the labeled activity in a long weakly labeled data sequence. In the future, we will utilize the location function of our model to collect specific segments extracted from the weakly labeled dataset as a dataset for human activity recognition, and then evaluate the performance.

\bibliographystyle{IEEEtran}
\bibliography{ref}

\end{document}